\documentclass[10pt,twocolumn,letterpaper]{article}

\usepackage[pagenumbers]{cvpr} 

\definecolor{cvprblue}{rgb}{0.21,0.49,0.74}
\usepackage[pagebackref,breaklinks,colorlinks,allcolors=cvprblue]{hyperref}
\usepackage{graphicx}	
\usepackage{hyperref}
\usepackage{url}
\usepackage{multirow}
\usepackage{makecell}
\usepackage{enumitem}

\def\confName{CVPR}
\def\confYear{2026}

\title{UniMo: Unifying 2D Video and 3D Human Motion with an Autoregressive Framework}

\author{Youxin Pang\textsuperscript{1,2}\footnotemark[1] \quad Yong Zhang\textsuperscript{2}\footnotemark[2] \quad Ruizhi Shao\textsuperscript{1} \quad Xiang Deng\textsuperscript{1,2}  \\ Feng Gao\textsuperscript{2} \quad Xiaoming Xu\textsuperscript{2} \quad Xiaoming Wei\textsuperscript{2} \quad Yebin Liu\textsuperscript{1}\footnotemark[2] \\ \textsuperscript{1}Tsinghua University\\
\textsuperscript{2}Meituan\\ \\
\href{https://carlyx.github.io/UniMo/}{https://carlyx.github.io/UniMo/} }
\begin{document}

\twocolumn[{
\maketitle
\begin{center}
    \captionsetup{type=figure}
    \includegraphics[width=1.\linewidth]{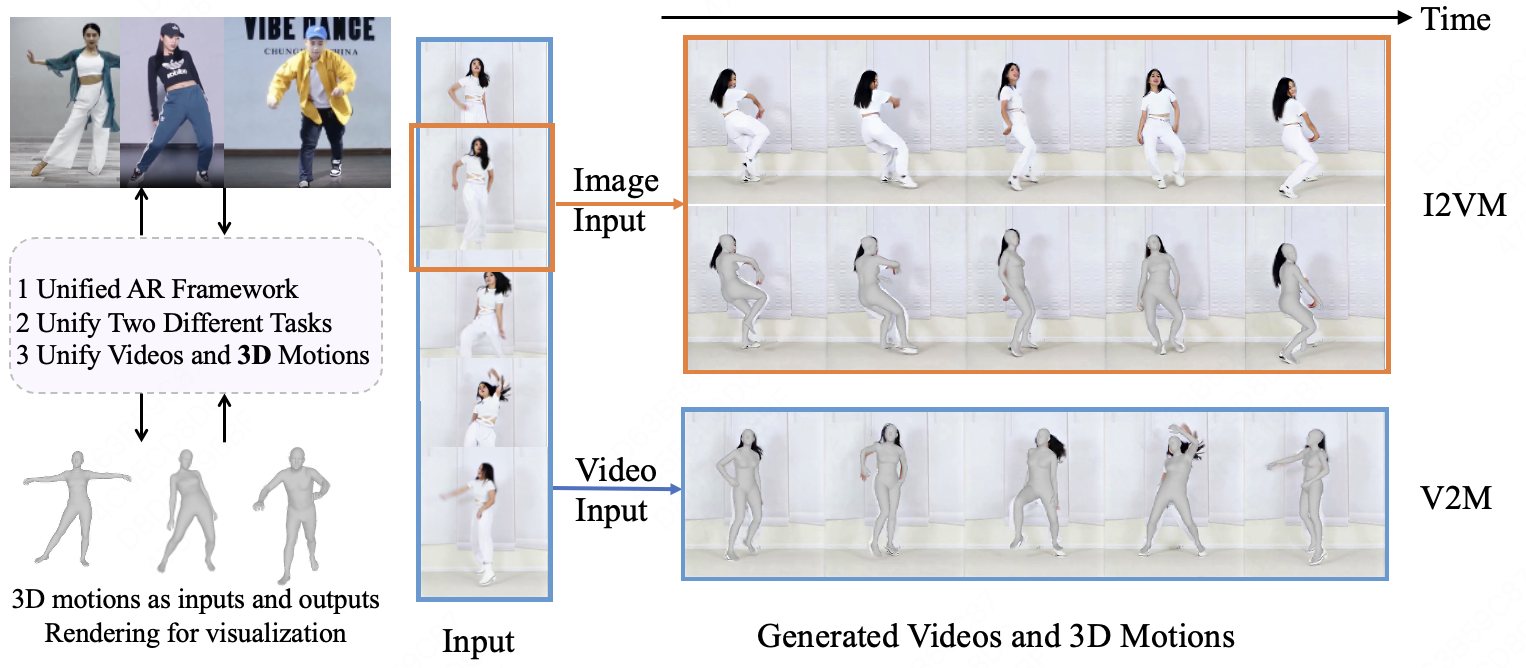}
    \caption{We present UniMo, an innovative autoregressive model for joint modeling of 2D human videos and 3D human motions within a unified framework. Left: Unlike existing methods that map 3D motions to 2D maps for video-motion alignment, we directly use 3D motions as inputs and outputs. Right: We unify the I2VM (Image-to-Video-and-Motion) and V2M (Video-to-Motion) tasks within a single transformer framework, demonstrating the effectiveness of the proposed method.
}
    \label{fig:teaser}
\end{center}
}]

\renewcommand{\thefootnote}{\fnsymbol{footnote}}
\footnotetext[1]{Work done during the internship at Meituan}
\footnotetext[2]{Corresponding authors.}

\begin{abstract}
We propose UniMo, an innovative autoregressive model for joint modeling of 2D human videos and 3D human motions within a unified framework, enabling simultaneous generation and understanding of these two modalities for the first time.
Current methods predominantly focus on generating one modality given another as the condition or integrating either of them with other modalities such as text and audio.
Unifying 2D videos and 3D motions for simultaneous optimization and generation remains largely unexplored, presenting significant challenges due to their substantial structural and distributional differences.
Inspired by the LLM's ability to unify different modalities, our method models videos and 3D motions as a unified tokens sequence, utilizing separate embedding layers to mitigate distribution gaps.
Additionally, we devise a sequence modeling strategy that integrates two distinct tasks within a single framework, proving the effectiveness of unified modeling.
Moreover, to efficiently align with visual tokens and preserve 3D spatial information, we design a novel 3D motion tokenizer with a temporal expansion strategy, using a single VQ-VAE to produce quantized motion tokens. 
It features multiple expert decoders that handle body shapes, translation, global orientation, and body poses for reliable 3D motion reconstruction.
Extensive experiments demonstrate that our method simultaneously generates corresponding videos and motions while performing accurate motion capture.
This work taps into the capacity of LLMs to fuse diverse data types, paving the way for integrating human-centric information into existing models and potentially enabling multimodal, controllable joint modeling of humans, objects, and scenes.
\end{abstract}    
\section{Introduction}
\label{sec:intro}
Digital human modeling is a fundamental task in computer vision.
Integrating 3D motions and 2D videos plays a pivotal role in extensive tasks, including human video synthesis and motion capture.
For human video synthesis, most methods~\cite{shao2024360, zhu2024champ, lin2025omnihuman, hu2024animate} focus on producing human videos that are consistent with input motions. 
In the realm of motion capture~\cite{goel2023humans, shen2024world, khirodkar2024sapiens}, they aim to capture the corresponding 3D motions from input videos.
However, the aforementioned methods primarily use one modality as the condition to generate another, without engaging in joint modeling and optimization of both modalities.

This paper focuses on unifying 2D human videos and 3D human motions to achieve joint optimization and generation.
Recently, large language models (LLMs)~\cite{brown2020language} have been widely applied in various large vision-language models~\cite{liu2023visual,bai2025qwen2,guo2025seed1,agarwal2025cosmos}, effectively capturing the relationships among different modalities such as audio, text, and vision.
Building on this, numerous approaches leverage LLM-style frameworks to integrate motions with different modalities, including vision~\cite{li2025chatmotion, chen2024motionllm, li2025unipose}, text~\cite{zhu2025motiongpt3}, and audio~\cite{luo2024m}.
While the above methods integrate multimodal data using LLM-style structure, they primarily focus on generating one modality based on another. 
The task of unifying 2D videos and 3D motions for simultaneous optimization and generation has not been thoroughly explored, especially in terms of aligning these two different modalities for joint generation.
Inspired by the LLM's ability to unify different modalities, we explore the possibility of transferring this capability to our specific task.


The primary challenge with 3D motions lies in the lack of explicit spatial correspondence with 2D videos, which inhibits integration via straightforward operations like addition or concatenation~\cite{hu2024animate}. 
In this paper, we introduce a novel autoregressive (AR) framework for joint modeling of 2D human videos and 3D human motions, achieving simultaneous optimization and generation of these two modalities for the first time.
Unlike existing single-task methods, our unified framework efficiently performs both generation and understanding tasks, as illustrated in Fig.~\ref{fig:teaser}, further advancing the ability of LLMs to integrate various data types.
For the generation task, the model simultaneously produces videos and corresponding 3D motions from a single image.
For the understanding task, it captures corresponding 3D motions from video inputs.
To unify these two tasks, we design a novel sequence modeling strategy that assigns corresponding tasks based on the input tokens.
Moreover, motion tokens differ from visual tokens with respect to token distribution, and the model simultaneously outputs tokens from two modalities, which can easily lead to potential confusion between them.
Therefore, we design distinct embedding layers for each modality to mitigate distribution gaps and enhance alignment, incorporating learnable vocabulary embeddings and positional embedding methods.

Another challenge lies in constructing 3D motion representations for seamless integration with visual information within our AR framework.
A straightforward approach is to represent 3D motions similarly to visual tokens. 
Recently,  MotionGPT~\cite{jiang2023motiongpt}, SOLAMI~\cite{jiang2025solami}, and Duolando~\cite{siyao2024duolando} have explored 3D motion tokenizers, based on 3D keypoints or SMPL(X) parameters. 
3D keypoints are relatively simple but insufficient to represent complex human motions.
Therefore, we use SMPL-X~\cite{pavlakos2019expressive} to model human motions, incorporating 3D parameters like body shapes, translation, global orientation, and body poses.
However, applying current tokenizers directly to our task introduces two main challenges.
Firstly, most methods employ temporal compression to reduce resource usage, which is effective for motion-text alignment but results in a token quantity imbalance in video-motion modeling.
Secondly, many approaches segment the human body into multiple parts, processing each separately with several VQ-VAEs.
Although this improves reconstruction accuracy, it results in multiple sets of motion tokens, thus increasing complexity for our unified model.
To overcome these challenges, we propose a novel 3D motion tokenizer that models all SMPL-X parameters using a single VQ-VAE, complemented by a novel temporal expansion strategy to enhance reconstruction accuracy and balance the quantity of vision tokens.
The proposed tokenizer generates motion tokens while ensuring accurate reconstruction, laying the foundation for effective multimodal fusion through AR.

By leveraging the proposed motion tokenizer and AR model, our approach consistently generates both videos and motions across two tasks, illustrating the potential of modeling 3D motions and 2D videos within a unified AR framework.
This work not only explores the capacity of LLMs to fuse diverse data types but also establishes a foundation for embedding human-centric information into existing architectures, potentially enabling multimodal, controllable joint modeling of humans, objects, and scenes.
We summarize our contributions as follows,
\begin{itemize}
    \item A novel LLM-style framework jointly models 3D human motions and 2D human videos, enabling simultaneous generation and optimization of both modalities for the first time.
    \item A novel 3D motion tokenizer employs a temporal expansion strategy to effectively quantize and reconstruct SMPL-X parameters, laying the foundation for modality integration and motion generation.
    \item A unified autoregressive model, featuring a novel sequence modeling and independent embedding strategy, integrates two distinct tasks within a single transformer model, effectively alleviating distribution gaps and aligning both modalities.
\end{itemize}
\section{Related Work}
\label{sec:related}

\noindent{\textbf{Human Video Synthesis.}} 
The objective is to generate a corresponding video given a single human image and a sequence of driving motion. 
Current approaches predominantly utilize diffusion-based methods to tackle this task, often employing UNet~\cite{hu2024animate, zhu2024champ, xu2024magicanimate,wang2024vividpose,chang2025x,kim2024tcan,zhang2024mimicmotion} or DiT~\cite{shao2024360,lin2025omnihuman,shao2025interspatial,ding2025mtvcrafter} structures. 
To incorporate the driving motions, they typically utilize 2D maps such as skeleton maps, normal maps, and densepose maps~\cite{karras2023dreampose}. 
However, they only align 2D motions with visual latents at the input stage and lack optimization of 3D motions.

\noindent{\textbf{Human Motion Capture.}}
Motion capture~\cite{shen2024world,goel2023humans,kanazawa2018end,rajasegaran2022tracking,kanazawa2019learning,kocabas2020vibe,luo20203d,khirodkar2024sapiens} is a classic task aiming at extracting corresponding human motions from video inputs.
For instance, HMR~\cite{kanazawa2018end} utilizes a CNN to regress SMPL parameters, while 4DHumans~\cite{goel2023humans} introduces a fully transformer-based model based on an enhanced HMR and 3D tracking system.
Besides, GVHMR~\cite{shen2024world} estimates human poses in a novel gravity-view coordinate to reduce ambiguity in image-pose mapping. 
Typically, they commence video preprocessing by tracking humans, detecting keypoints, and extracting features, followed by regressing motion parameters from these features. 
Besides, they primarily focus on the transfer from videos to motions, lacking emphasis on the joint modeling of the two modalities.

\noindent{\textbf{Human Video-Motion Joint Tasks.}}
Some methods~\cite{li2025chatmotion, chen2024motionllm, li2025unipose} employ joint modeling of videos and motions to enhance the understanding of human behavior.
However, these approaches merely integrate the two modalities into a unified representation at the input stage, lacking the capability to simultaneously generate both.
VideoJAM~\cite{chefer2025videojam}, AnimaX~\cite{huang2025animax}, and OmniVDiff~\cite{xi2025omnivdiff} introduce an appearance-motion aligning framework based on diffusion models, which represent motion information using 2D motion maps. 
While 2D motion maps are flexible and can be easily represented as RGB videos, they inevitably suffer from the loss of crucial 3D spatial information.
SViMo~\cite{dang2025svimo} adopt diffusion frameworks to simultaneously generate hand-object videos and motions.


\noindent{\textbf{Human 3D Motion Tokenizer.}}
The human motion tokenizer~\cite{ding2025mtvcrafter} is designed to compress and convert raw motion data, such as keypoints and SMPL-X parameters, into motion tokens. MotionGPT~\cite{jiang2023motiongpt} and SOLAMI~\cite{jiang2025solami} pre-train the 3D human motion tokenizers using the VQ-VAE architecture.
However, their tokenizers require complex and extensive data processing, including orientation adjustments, foot contact modifications, and the application of forward/inverse kinematics, which demands predefined human kinematic chains tailored to specific datasets.
Recently, Duolando~\cite{siyao2024duolando} introduces a simplified motion tokenizer that uses raw 3D joint coordinates as inputs. 
Nevertheless, they independently process body poses and translation, and 3D joints remain insufficient to fully capture complex human motions.

\noindent{\textbf{LLM-style Video and Motion Models.}}
Recently, autoregressive models~\cite{wu2025janus, wang2024emu3,chen2025janus, wu2024deepseek} based on transformer architectures have demonstrated impressive results in multimodal modeling. 
Cosmos~\cite{agarwal2025cosmos} approaches world simulation generation as a next-token prediction task, akin to language modeling, and incorporates text embeddings using cross-attention.
Additionally, many methods~\cite{jiang2023motiongpt, jiang2025solami, chen2024motionllm, li2025human} leverage LLMs to achieve unified generation and understanding of motions alongside multiple modalities such as text and audio.
\section{Method}\label{sec:met}
\begin{figure*}[tp]
    \centering
    \includegraphics[width=1.\linewidth]{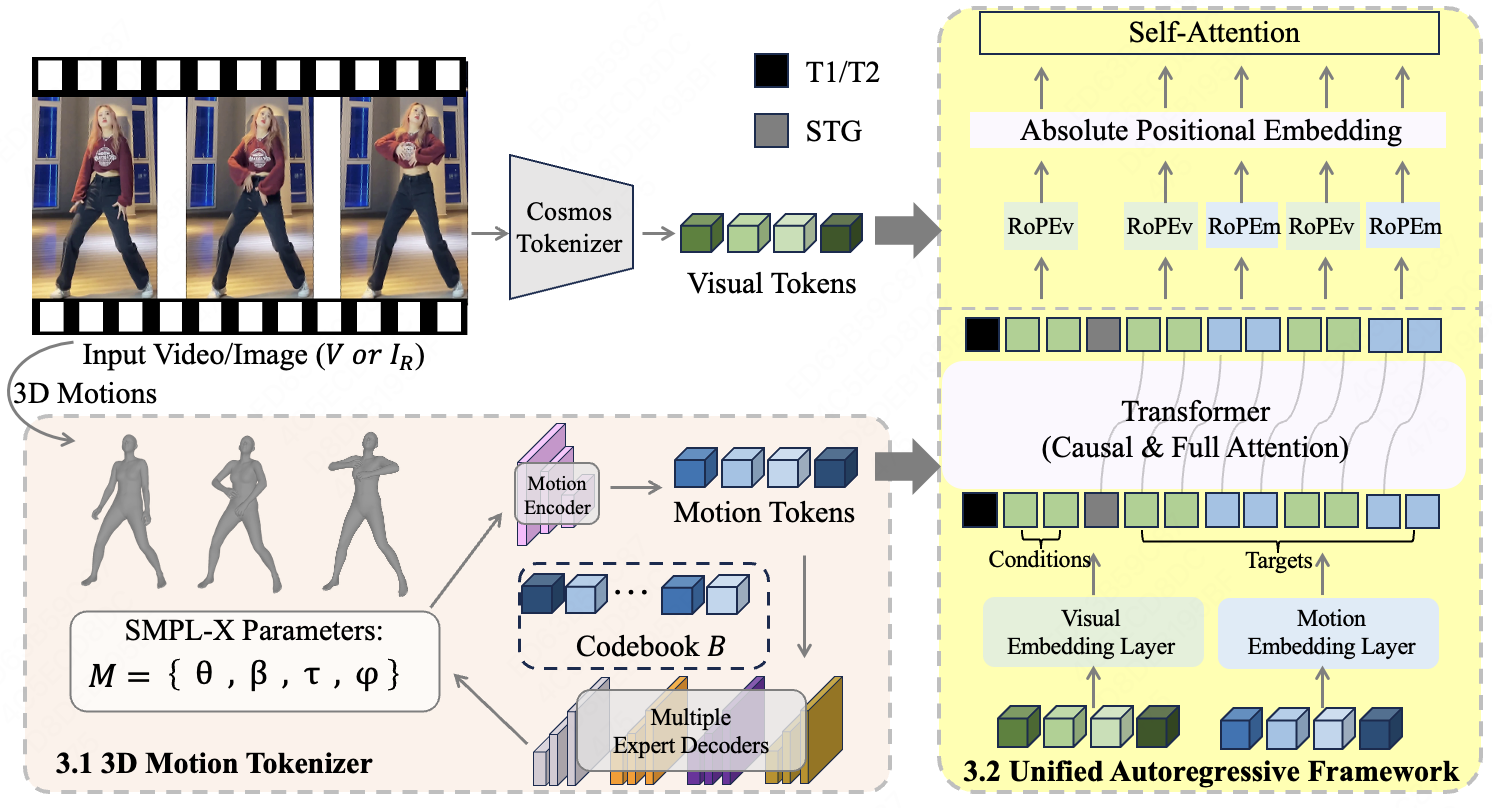}
    \caption{Overview of UniMo. Left: We introduce a 3D motion tokenizer that is responsible for quantizing raw 3D motions $M$ into motion tokens corresponding to visual tokens and accurately reconstructing back to $M$ from these tokens. The 3D motion tokenizer comprises a motion encoder, a learnable codebook $B$, and multiple expert decoders. Right: Given visual tokens and motion tokens, we propose an AR transformer framework with new sequence modeling strategies to unify the two modalities, enabling the execution of two distinct tasks (only the I2VM task is illustrated in the figure). To better integrate the two types of tokens, we propose independent vocabulary embedding layers and different positional embedding ways.}
    \label{fig:pipeline}
\end{figure*}

The overall pipeline of UniMo is illustrated in Fig.~\ref{fig:pipeline}. 
Our goal is to model 3D motions and 2D videos without relying on 2D motion maps, achieving simultaneous generation and optimization of them.
Specifically, UniMo employs a unified AR framework to integrate the two modalities and implement two distinct tasks for validating the generation and understanding capabilities.
Notably, conventional understanding tasks are defined as understanding visual information and generating textual responses based on queries. 
In contrast, our task focuses on understanding human information from the video and capturing the corresponding 3D motion.
In the image-to-video-and-motion task (I2VM), given a single reference image $I_{R}$, the goal is to generate subsequent $T$-frame videos along with corresponding 3D motions $M_{k=1}^{T}$.
In the video-to-motion task (V2M), given a video sequence $V_{k=1}^{T}$, the objective is to capture the corresponding 3D motions.

We first introduce a 3D motion tokenizer tailored for our task in Sec.~\ref{sec:3dmt}, which is responsible for quantizing raw 3D motions $M$ into motion tokens corresponding to visual tokens. 
Additionally, this tokenizer can accurately reconstruct $M$ from the quantized tokens, establishing a foundation for joint modeling.
Then, we detail how autoregressive modeling can unify visual and motion tokens within a transformer architecture in Sec.~\ref{sec:uaf}, encompassing task-specific sequence modeling strategies and independent embeddings adapted to various modalities.
By simultaneously optimizing two modalities and tasks, the model enhances their correspondence and improves generative and understanding capabilities.
After that, we discuss the training strategies utilized for multi-task unification in Sec.~\ref{sec:tis}.


\subsection{3D Motion Tokenizer}
\label{sec:3dmt}
To integrate 3D motions $M$ with visual tokens and enable their generation, it is necessary to develop a tokenizer that accepts $M$ as input, quantifying them into discrete tokens akin to visual tokens, while also being capable of reconstructing the original $M$. 
MotionGPT~\cite{jiang2023motiongpt}, SOLAMI~\cite{jiang2025solami}, and Duolando~\cite{siyao2024duolando} have explored 3D motion tokenizers for this target, based on 3D keypoints or SMPL(X) parameters. 
3D keypoints are relatively simple but insufficient to fully represent complex 3D human motion information.
Moreover, obtaining SMPL(X) parameters offers greater value in motion generation tasks.
MotionGPT and SOLAMI are SMPL(X)-based methods that require complex processing of motions, including orientation adjustments, foot contact modifications, and the application of forward/inverse kinematics, making it difficult to generalize across diverse datasets.
Additionally, SOLAMI divides the human body into multiple parts and designs several VQVAEs to independently learn different segments and translations. 
This configuration generates multiple sets of tokens, increasing complexity in our unified model.


To address this, as shown in the left box in Fig.~\ref{fig:pipeline}, we use SMPL-X models~\cite{pavlakos2019expressive} to comprehensively represent 3D human motions and propose a novel 3D motion tokenizer tailored to our task.
Specifically, SMPL-X is a unified body model with shape parameters trained jointly for the face, hands and body.
In our task, the body part is used to represent human motions through parameterized body poses ($\theta \in \mathbb{R}^{T \times 63}$), shape coefficients ($\beta \in \mathbb{R}^{T \times 10}$), global orientation ($\phi \in \mathbb{R}^{T \times 3}$), and translation ($\tau \in \mathbb{R}^{T \times 3}$).
The proposed tokenizer is capable of taking the entire set of $M=(\theta, \beta, \phi, \tau)$ as inputs, quantizing them into discrete tokens, and accurately reconstructing the SMPL-X parameters from motion tokens.
Inspired by Duolando~\cite{siyao2024duolando}, we employ a VQ-VAE structure comprising an encoder, a learnable codebook, and multiple expert decoders. 
We first cascade $M$ along the last dimension channel, which result in channels $C = (63+10+3+3)$.
Then the encoder uses 1D convolutions transforming them into high-dimensional semantic features $F \in \mathbb{R}^{T' \times C}$, where $T' = T / s $ and $s$ is the scaling factor. 
In data processing, the absolute position of the first frame is preserved, while subsequent frames are transformed into velocity representations by subtracting the position of the preceding frame, thereby reinforcing temporal continuity.
During reconstruction, the original positions can be restored using prefix sum techniques.
Subsequently, the sequence $F_{k=1}^{T'} = \{f_{1}, f_{2}, ..., f_{T'}\}$ is quantized by replacing each $f_{k}$ with the nearest element in the codebook $B$, transforming it into a discrete sequence of tokens.
Finally, we employ four expert decoders consisting of 1D convolutions to individually reconstruct parameters $\theta$, $\beta$, $\tau$, and $\phi$ from tokens.

Notably, most methods employ temporal compression to reduce resource usage, which is effective for motion-text alignment but results in a token quantity imbalance in our task.
Besides, SMPL-X parameters are more complex than 3D keypoints, resulting in greater learning difficulty and poorer performance on motion metrics compared to keypoints~\cite{jiang2025solami}.
To address this issue, we adopt an expansion strategy for temporal processing. 
By setting $s=1/36$, we represent the SMPL-X parameters for one frame with 36 discrete tokens. 
This serves two purposes: (1) Visual tokens significantly outnumber motion tokens, thus expanding motion tokens helps balance the disparity in their quantities to some extent; (2) One of our objectives is the accurate reconstruction of SMPL-X parameters, and expanding tokens can improve accuracy.
Please refer to the supplementary materials for the analysis of the impact of the quantity of tokens on computational cost.

\begin{figure*}[tp]
    \centering
    \includegraphics[width=1.\linewidth]{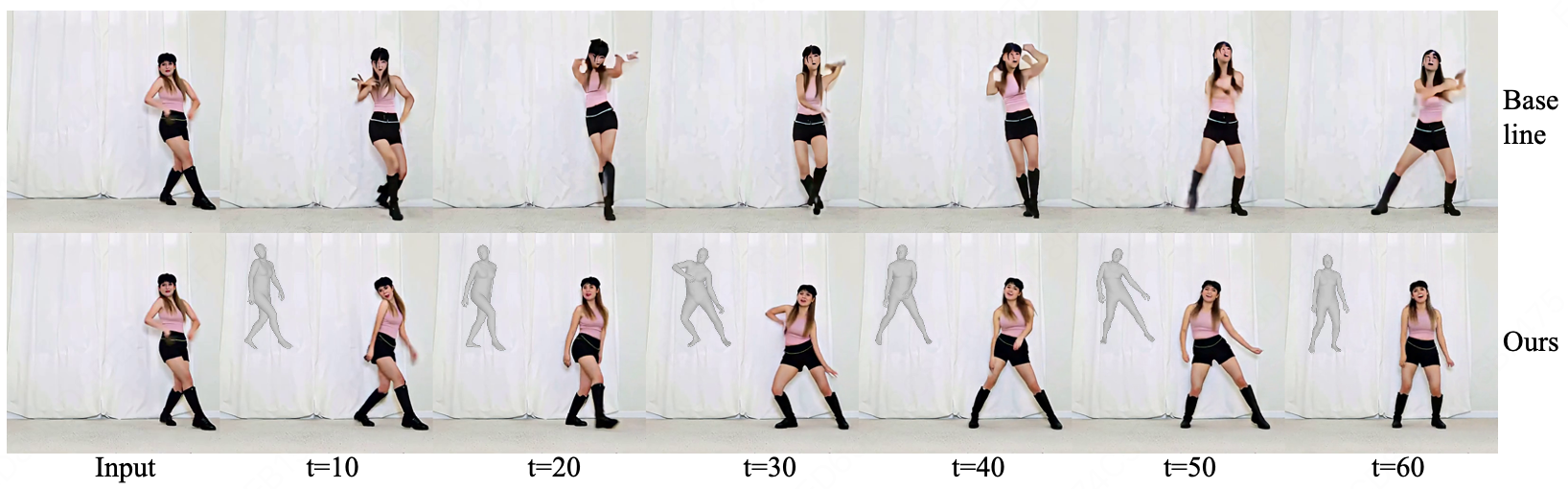}
    \caption{Comparison with the baseline on I2VM tasks. We present results in temporal order from left to right, sampling one frame for every 10 generated frames.
    In our results, the simultaneously generated 3D motions are rendered and visualized in the top-left corner, while the baseline does not generate motion.
    For a fair comparison, the baseline is finetuned using our training video dataset.
    }
    \label{fig:i2vm_comp}
\end{figure*}

\subsection{Unified Autoregressive Framework}
\label{sec:uaf}
\textbf{Unified Motion-Visual Representation.}
Building on the impressive performance of Cosmos~\cite{agarwal2025cosmos} in autoregressive (AR) video modeling, we adopt the Cosmos AR framework as our backbone model.
We utilize the Cosmos tokenizer to quantize and compress the videos into visual tokens with a compression rate of $8 \times 16 \times16$.
However, in Cosmos, multimodal sequence modeling is accomplished through cross-attention, which prevents simultaneous generation and optimization of another modality.
Inspired by LLMs~\cite{xie2025show, wang2024emu3}, we structure 3D motion tokens and visual tokens into a unified sequence following an interleaved motion-video format.
Given that our framework involves I2VM and V2M tasks, we introduce special tokens to identify different tasks.
For V2M task, we format the sequence as:
\begin{equation}
    [T1] \ [Vt_{1} \ Vt_{2} \ ... \ Vt_{N}] \ [STG] \ [Mt_{1} \ Mt_{2} \ Mt_{M}],
\end{equation}
where $T1$ means V2M task. 
$Vt$ and $Mt$ represent the visual and motion tokens, respectively.
$STG$ marks the beginning of generation, with the conditional sequence placed before it and the target sequence positioned after it.
$N$ and $M$ represent the number of visual and motion tokens, respectively.
Notably, we use 16 frames as a unit, meaning each $Vt$ or $Mt$ represents information from 16 frames.

For I2VM task, we format the sequence as:
\begin{equation}
    [T2] \ [It] \ [STG] \ [Vt_{1}] \ [Mt_{1}] \ [Vt_{2}] \ [Mt_{2}] \ ... \ [Vt_{N}] \ [Mt_{M}],
    \label{eq:i2vm}
\end{equation}
where $T2$ means I2VM task. $It$ is the single reference image tokens.
The sequence formats above are flexible by using a task-specific token at the beginning to distinguish two tasks, and employing the $STG$ special token to separate the conditions from the targets within different tasks.
Notably, in Eq.~\ref{eq:i2vm}, we define the target sequence as interleaved visual tokens and motion tokens.
This design strengthens the model's capability to integrate both modalities simultaneously, enabling the generation process to leverage all previously incorporated modalities.

\textbf{Vocabulary Embedding Layers.}
In the Cosmos AR model, sequence modeling involves only visual modality, and thus uses a single embedding layer to process all discrete tokens. 
However, due to the inherent gaps between motion and visual tokens, using a single embedding layer may lead to distribution entanglement. 
Additionally, simultaneous generation of two modalities within one framework raises the possibility of output modality confusion. 
To address these issues, as shown in the right box in Fig.~\ref{fig:pipeline}, we employ two separated learnable embedding layers, one for visual tokens and the other for motion tokens.

\textbf{Positional Embedding Layers.}
Similar to Cosmos AR model~\cite{agarwal2025cosmos}, we use Rotary Position Embedding (RoPE) and Absolute Positional Embedding (APE) to model positional embeddings.
Firstly, we apply APE to the entire sequence to directly model the positional relationship of interleaved modalities. 
Additionally, to better align them, we implement two different RoPE modeling approaches to independently process each modality (right box in Fig.~\ref{fig:pipeline}).
Specifically, for video embeddings, we adopt 3D factorized RoPE to model both temporal and spatial dimensions.
Given the limited joint spatial information in motion embeddings, we apply RoPE to the temporal dimension.
Finally, we achieve modality alignment by aligning the positions of two modalities in RoPE.
Specifically, for queries ($Q$) and keys ($K$) in attention operations:
\begin{align}
    \hat{Q} &= (RoPE_{m}(Q_{m}) \oplus RoPE_{v}(Q_{v})) + APE(Q_{e}), \\
    \hat{K} &= (RoPE_{m}(K_{m}) \oplus RoPE_{v}(K_{v})) + APE(K_{e}),
\end{align}
where $Q_{e}$ denotes the entire sequence, $\hat{Q}$ represents the results after applying positional embedding. $RoPE_{m}$ and $RoPE_{v}$ represent motion RoPE and visual RoPE respectively. $Q_{m}$ and $Q_{v}$ represent motion tokens and visual tokens respectively, and $\oplus$ denotes concatenation operations. ($K$ is processed similarly)

\subsection{Training Strategy}
\label{sec:tis}
UniMo is trained with a two-stage training approach. 
In the first stage, we train the 3D motion tokenizer in an end-to-end way using 3D motion data, and the training loss is:
\begin{equation}
    \mathcal{L}_{VAE} = \mathcal{L}_{rec}(M', M_{gt}) + \lambda\|F - sg(B)\| + \|sg(F) - B \|,
    \label{eq:vqloss}
\end{equation}
where $\mathcal{L}_{rec}$ is $l_{1}$ loss between the predicted value $M'$ and the real value $M_{gt}$. $sg(\cdot)$ means stop
gradient operation and $\lambda$ is the trade-off parameter.
Similar to Duolando, we add velocity and acceleration to perform $\mathcal{L}_{rec}$.
Notably, the motion encoder is used only during the training phase.

In the second stage, the parameters of tokenizer are frozen to serve as a quantizer, supplying discrete motion tokens for training the AR model.
Attention masks are crucial for AR models training, with most methods utilizing causal masks to ensure that the current token can only attend to preceding tokens. 
In our task, we apply causal masks to the target sequence, while employing full masks on the conditional sequence to enhance bidirectional context awareness.
In addition, we combine the data from the two tasks in equal proportions to unify the multimodal and multi-task, and train the AR model in an end-to-end way.
The loss is the cross-entropy:
\begin{equation}
    \mathcal{L}_{AR} = - \sum_{i=1}^{L} \log p (q_{i} | q_{<i}, c),
\end{equation}
where $L$ is target sequence length. $q_{i}$ is the i-th token in the sequence and $c$ means the conditions.

\section{Experiments}\label{sec:expe}

\begin{table}[tp]
\caption{Evaluation of the 3D motion tokenizer. For a fair comparison, we train SOLAMI with Human4DiT-Video dataset and remove the hand VQ-VAE.}
\centering
\label{tab:token_comp}
\scalebox{0.9}{
\begin{tabular}{l|llll}
\hline
{ Methods }  & { MPJPE $\downarrow$} & { PA-MPJPE$\downarrow$} & { PVE$\downarrow$} & { Accel$\downarrow$} 
\\ 
\hline
SOLAMI & 24.3354 &  14.7212 &  29.6462 & 7.7384 \\ 
Our &  \bf 8.6344 &  \bf 5.3876 &  \bf 10.7010 &  \bf 2.4632 \\
\hline
\end{tabular}
}
\end{table}

\begin{figure*}[tp]
    \centering
    \includegraphics[width=1.\linewidth]{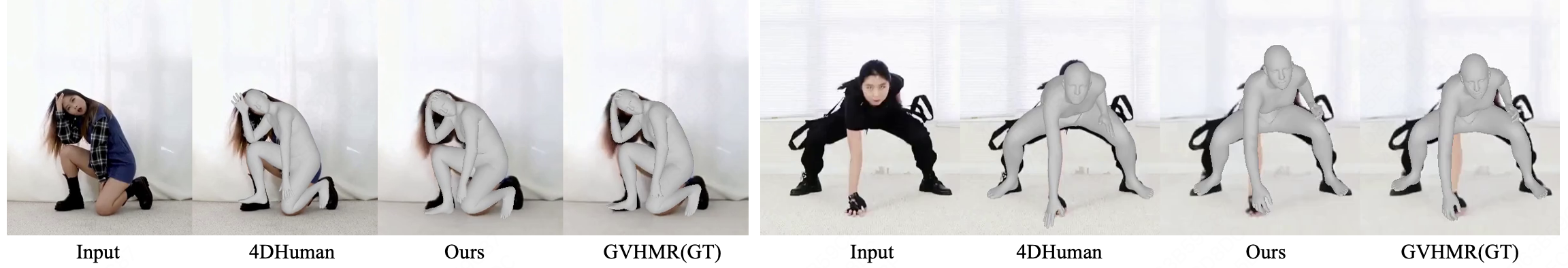}
    \caption{Comparison with different methods on the V2M task.
    It demonstrates that our approach achieves results comparable to current state-of-the-art methods.}
    \label{fig:comp_v2m}
\end{figure*}

\begin{table*}[tp]
\caption{Comparison of I2VM tasks. For motion-video consistency, GVHMR is used to extract motions as pseudo ground truth. Motion diversity is assessed using the metrics from Duolando. The generated videos are evaluated with VBench, focusing on appearance (App.) and visual motion (Mot.). The baseline (Cosmos) lacks the ability to generate motion and doesn't involve consistency.
}
\centering
\label{tab:i2vm_motoin}
\begin{tabular}{l|llll|ll|ll}
\hline
\multirow{2}*{Method}  & \multicolumn{4}{c|}{\makecell[c]{Video-Motion Consistency\\}}  & \multicolumn{2}{c|}{\makecell[c]{Motion Diversity\\}} & \multicolumn{2}{c}{\makecell[c]{Video Quality\\}} \\ & { MPJPE $\downarrow$} & { PA-MPJPE$\downarrow$} & { PVE$\downarrow$} & { Accel$\downarrow$} & { FID$\downarrow$} & { DIV$\uparrow$} & { App.$\uparrow$} & { Mot.$\uparrow$}
\\ 
\hline
Baseline & - & - & - & - & 35.7305 & 10.8601 & 0.7743 & 0.9249 \\ 
Ours & 41.3058 & 30.9548 & 47.3111 & 3.9984 & \bf 27.3984 & \bf 12.2522 & \bf 0.8516 & \bf 0.9441 \\ 
\hline
\end{tabular}
\end{table*}

\subsection{Settings}
\textbf{Metrics.} We employ distinct metrics to evaluate the generated videos and motions.
For video evaluation, similar to VideoJAM~\cite{chefer2025videojam}, we utilize VBench~\cite{zheng2025vbench} to analyze various disentangled features, including appearance and visual motion attributes.
Please refer to our supplementary paper for more results.
We evaluate the generated 3D motions in two ways.
For I2VM tasks, we use FID and diversity (DIV) metrics to compare the distribution of generated motions with ground-truth motions.
In line with Duolando~\cite{siyao2024duolando}, we derive FID and diversity from motion features in AIST++~\cite{li2021ai}.
For V2M tasks, following WHAM~\cite{shin2024wham}, we report metrics such as MPJPE, PA-MPJPE, PVE, and Accel.

We train our model on Human4DiT-Video~\cite{shao2024360}, a dataset comprising 10k in-the-wild monocular video clips with corresponding motion sequences.  
Besides, we observe temporal jitter in the motion dataset within Human4DiT-Video, prompting us to use GVHMR~\cite{shen2024world} to re-extract the 3D motion data. 
For evaluation, we select 300 single-human clips from Human4DiT-Video~\cite{shao2024360}, DNA-Rendering~\cite{cheng2023dna}, 3DPW~\cite{vonMarcard2018}, RICH~\cite{Huang:CVPR:2022}, and BEDLAM~\cite{black2023bedlam} as the testset. 
It is important to note that our model actually uses 3D motions (SMPL-X) as both inputs and outputs. 
For intuitive visualization, we render the generated 3D motions into 2D maps in all visual results.
In addition, due to the poor visual effect of video tokenizer in Cosmos, we use SeedVR~\cite{wang2025seedvr} to enhance the visual quality of generated videos, which does not disrupt our core goal of joint modeling the two modalities.
Please refer to our supplementary paper for more implementation details.

\subsection{Baseline Comparisons}
\textbf{3D Motion Tokenizer.}
Unlike existing methods such as SOLAMI, which processes the human body into parts with multiple VQ-VAEs, we utilize a single VQ-VAE to process the entire SMPL-X parameters.
To ensure the reconstruction accuracy, we propose a temporal expansion strategy.
Specifically, for a fair comparison, we use Human4DiT-Video to fine-tune SOLAMI. Additionally, since our task does not involve hand modeling, we remove the VQVAE for hands from SOLAMI.
As shown in Tab.~\ref{tab:token_comp}, our tokenizer exhibits superior performance in reconstruction accuracy. 
Please refer to the supplementary paper for more results.

\textbf{Image-to-Video-Motion Task.}
Since our method is based on an AR structure, we employ Cosmos~\cite{agarwal2025cosmos} as the baseline for comparison, which adopts the same architecture with ours.
Specifically, Cosmos is a video generation model. 
To ensure a fair comparison, we fine-tune it on videos from our training dataset for the image-to-video task.
As illustrated in Fig.~\ref{fig:i2vm_comp}, when provided with a single image input, the baseline tends to generate distorted and physically implausible limbs, whereas our method produces more realistic movements.
This may be because training solely on human videos lacks an understanding of limb relationships, while incorporating motion enhances the realistic constraints on video generation. 
In the quantitative comparisons, we use VBench~\cite{zheng2025vbench} to evaluate the generated videos, focusing on appearance (App.) and visual motion (Mot.) aspects.
Additionally, the generated motions are evaluated from two dimensions: consistency with the generated videos and motion diversity.
For motion-video consistency, we use GVHMR to extract motions from the generated videos as pseudo ground truth. 
As the baseline model, Cosmos lacks motion generation capabilities, preventing us from evaluating consistency.
For motion diversity, we extract motions from generated videos from Cosmos using GVHMR to represent its motion outputs.
The quantitative results, shown in Tab.~\ref{tab:i2vm_motoin}, are consistent with the visual results.
Our predicted motions align well with the generated video.
Although the baseline and our method are relatively close in terms of motion diversity and visual dynamics, the artifacts generated by Cosmos are quite noticeable, resulting in reduced visual quality.
Please refer to the supplementary materials for more results.

\begin{figure*}[tp]
    \centering
    \includegraphics[width=1.\linewidth]{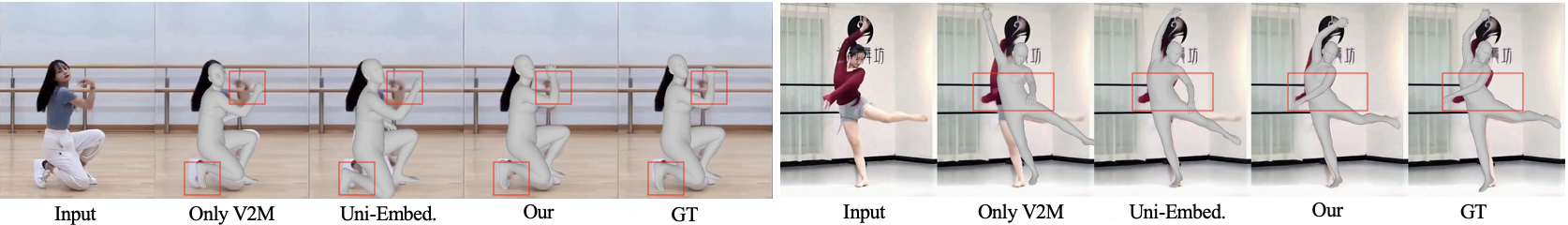}
    \caption{The ablation experiment on the V2M task. We compare our method against single-task training (Only V2M) and the approach without independent embedding (Uni-Embed.).}
    \label{fig:abl_v2m}
\end{figure*}

\textbf{Video-to-Motion Task.}
To demonstrate the potential of joint modeling, we validate the effectiveness of the V2M task across three datasets: 3DPW, RICH, and Human4DiT-Video.
Notably, as our primary focus is on exploring the possibility of joint modeling 3D motion and visual information, our experiments are conducted on single-person cases. 
Accordingly, we select single-person videos to construct the testset.
As illustrated in Fig.~\ref{fig:comp_v2m} and Tab.~\ref{tab:v2m_comp}, our method achieves results comparable to current state-of-the-art methods.
For more quantitative comparisons and results, please refer to the supplementary paper.

\begin{table}[tp]
\caption{Evaluation of the V2M task on Human4DiT-Video dataset. Since we use SMPL-X extracted by GVHMR as pseudo ground truth, GVHMR is excluded from metric calculation. }
\centering
\label{tab:v2m_comp}
\scalebox{0.9}{
\begin{tabular}{l|llll}
\hline
{ Methods }  & { MPJPE $\downarrow$} & { PA-MPJPE$\downarrow$} & { PVE$\downarrow$} & { Accel$\downarrow$} 
\\ 
\hline
GVHMR & - &  - &  - & - \\ 
4DHuman & 56.0701 &  35.4769 &  67.7157 & 15.7393 \\ 
Our &  \bf 43.2689 & \bf 28.1528 &  \bf 52.0143 & \bf 4.5647 \\ 
\hline
\end{tabular}
}
\end{table}

\begin{table}[tp]
\caption{Evaluation of our 3D motion tokenizer on different settings. For 
codebook utilization (Code. Util.), we randomly select 100 samples from the testset and calculate the proportion of different tokens, represented as $n/B$, where $n$ represents the number of different tokens.}
\centering
\label{tab:sup_token}
\scalebox{0.9}{
\begin{tabular}{l|llllll}
\hline
{ Methods }  & { MPJPE $\downarrow$} & { PVE$\downarrow$} & { Accel$\downarrow$} & {Code. Util.} 
\\ 
\hline
s=1/1 B=512 & 136.6638 &  155.1556 & 6.4386 &  80.08\% \\ 
s=1/8 B=512 &  69.1781 &    97.2101 &  5.3642 & 91.99\% \\ 
s=1/24 B=256 &  52.9926 &    85.0266 &  4.5206 & 97.66\% \\
s=1/24 B=512 &  10.4653 &    12.7147 &  2.7336 &  96.09\% \\
s=1/24 B=1024 &  47.4062 &    53.6114 &  4.5009 &  78.52\% \\
s=1/36 B=256 &  26.8738 &    35.1941 &  3.1909 &  99.61\% \\
s=1/36 B=512 &  \bf 8.6344  &  \bf 10.7010 &  \bf 2.4632 &  98.83\%\\
s=1/36 B=1024 & 30.8471 &    44.2118 &  3.7807 &  86.13\% \\
s=1/54 B=512 &  11.8932 &    17.878 &  2.9688 &  99.80\% \\
\hline
\end{tabular}
}
\vspace{-0.5 cm}
\end{table}
\subsection{Ablation Studies}
\textbf{3D Motion Tokenizer.}
To improve reconstruction accuracy and balance the number of video tokens, we propose expanding the motion token representation by using 36 tokens for each frame's SMPL-X parameters $M$.
Specifically, our experiments utilize GVHMR to extract the motions as pseudo ground truth.
We conduct experiments to thoroughly explore the effects of varying $B$ and $s$.
As illustrated in Tab.~\ref{tab:sup_token}, increasing the codebook size results in decreased utilization and a decline in reconstruction accuracy.
This phenomenon may occur because 3D motion is relatively simple, allowing a relatively small $B$ to sufficiently represent most distributions.
However, when $B$ becomes too large, it increases the degree of overfitting.
Furthermore, when $s$ is large (such as 1/1), the representation struggles to capture complex motions effectively.
Conversely, a smaller $s$ (such as 1/54) increases codebook utilization by providing more tokens to represent motions.
Despite this, reconstruction accuracy may suffer due to overfitting, suggesting that a single frame of motion does not require an excessive number of tokens for accurate representation.
Through quantitative and qualitative comparisons, we demonstrate that our 3D motion tokenizer effectively quantizes and reconstructs $M$, thereby establishing a solid foundation for AR training.
Please refer to the supplementary materials for an analysis on how token quantity impacts computational cost.

\begin{figure}[tp]
    \centering
    \includegraphics[width=1.\linewidth]{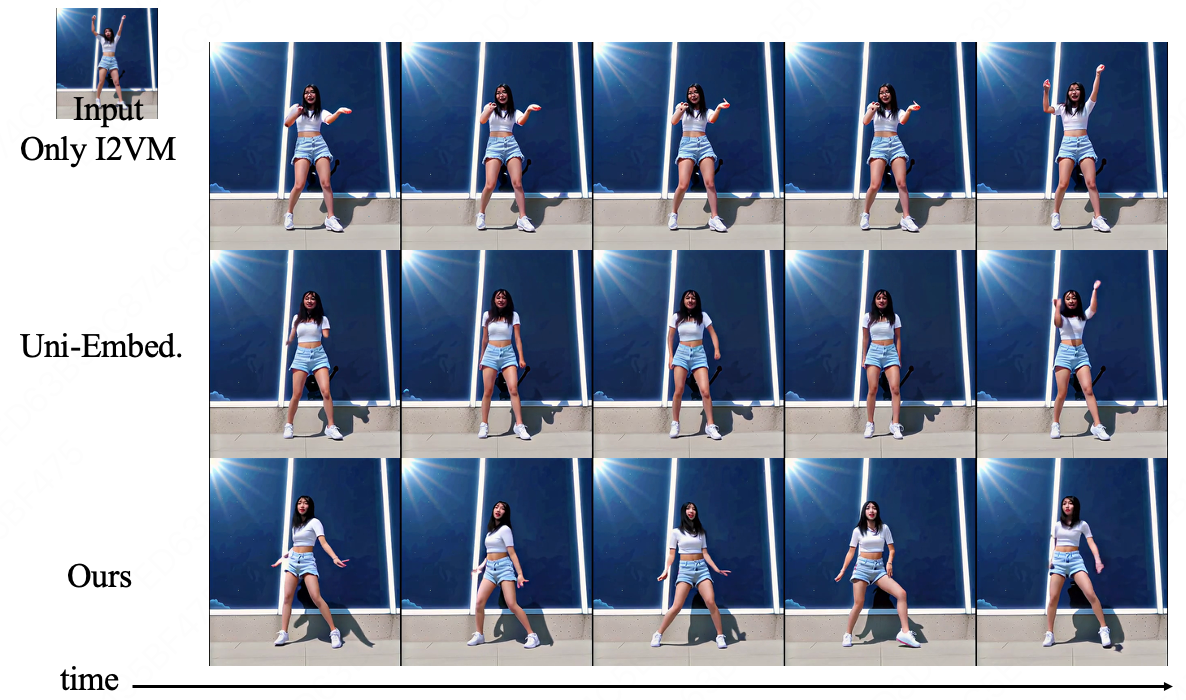}
    \caption{The ablation experiment on the I2VM task. We compare our method with single-task training (Only I2VM) and the approach without independent embedding (Uni-Embed.). The reference image is in the top left corner.}
    \label{fig:sup_abl_i2vm}
\end{figure}

\textbf{Independent Embeddings.}
We employ independent embeddings for discrete tokens and positions to mitigate distribution disparities between the two modalities and align them, thereby avoiding confusion when simultaneously outputting video and motion tokens within the same transformer model.
As illustrated in Fig.~\ref{fig:abl_v2m} and Fig.~\ref{fig:sup_abl_i2vm}, we explore the use of a single embedding layer for both visual and motion tokens, alongside a unified RoPE embedding for the entire sequence.
We then train the unified embedding (Uni-Embed) jointly on two tasks.
It is evident that for complex motions, Uni-Embed is less effective in capturing local details, such as those involved in squats, and struggles to generate a variety of movements.
Please refer to our supplementary paper for more details.

\textbf{Single Task.}
We integrate two tasks, I2VM and V2M, within the same transformer framework to achieve two objectives. 
First, the two tasks are used to validate the effectiveness of the proposed method. 
Second, we observe that training two tasks together yields better results than training each one individually, indicating a synergistic effect between the two tasks.
As illustrated in Fig.~\ref{fig:abl_v2m} and Fig.~\ref{fig:sup_abl_i2vm}, independent training of the V2M task fails to attain enhanced precision.
Additionally, training only the I2VM model leads to a decrease in motion diversity.
Please refer to our supplementary paper for more qualitative and quantitative results.
\section{Conclusion}
\label{sec:conclusion}

In this work, we introduce an innovative autoregressive model for the joint modeling of 2D human videos and 3D human motions within a unified framework. 
We propose a novel 3D motion tokenizer to establish a direct connection between 3D motions and visual information, thereby avoiding the use of 2D motion maps as the proxy.
By designing task-specific sequence modeling strategies and independent embedding methods, the proposed approach effectively integrates the two modalities and tasks.
Our extensive experiments show that the model generates corresponding videos and motions while capturing accurate motions, proving the potential of the proposed method and paving the way for advanced 3D/4D human modeling.
{
    \small
    \bibliographystyle{ieeenat_fullname}
    \bibliography{main}

@String(CVPR= {IEEE Conf. Comput. Vis. Pattern Recog.})

@String(ICCV= {Int. Conf. Comput. Vis.})

@String(ECCV= {Eur. Conf. Comput. Vis.})

@String(NIPS= {Adv. Neural Inform. Process. Syst.})

@String(TOG= {ACM Trans. Graph.})

@String(ACCV  = {ACCV})

@String(ICLR = {Int. Conf. Learn. Represent.})

@String(CVPR  = {CVPR})

@String(ICCV  = {ICCV})

@String(ECCV  = {ECCV})

@String(NIPS  = {NeurIPS})

@String(TOG   = {ACM TOG})

@String(ICLR  = {ICLR})

@article{shao2024360,
  title={360-degree human video generation with 4d diffusion transformer},
  author={Shao, Ruizhi and Pang, Youxin and Zheng, Zerong and Sun, Jingxiang and Liu, Yebin},
  journal=TOG,
  volume={43},
  number={6},
  pages={1--13},
  year={2024},
  publisher={ACM New York, NY, USA}
}

@inproceedings{zhu2024champ,
  title={Champ: Controllable and consistent human image animation with 3d parametric guidance},
  author={Zhu, Shenhao and Chen, Junming Leo and Dai, Zuozhuo and Dong, Zilong and Xu, Yinghui and Cao, Xun and Yao, Yao and Zhu, Hao and Zhu, Siyu},
  booktitle=ECCV,
  pages={145--162},
  year={2024},
  organization={Springer}
}

@article{lin2025omnihuman,
  title={Omnihuman-1: Rethinking the scaling-up of one-stage conditioned human animation models},
  author={Lin, Gaojie and Jiang, Jianwen and Yang, Jiaqi and Zheng, Zerong and Liang, Chao},
  journal={arXiv preprint arXiv:2502.01061},
  year={2025}
}

@article{hu2024animate,
  title={Animate anyone: Consistent and controllable image-to-video synthesis for character animation},
  author={Hu, Li and Gao, Xin and Zhang, Peng and Sun, Ke and Zhang, Bang and Bo, Liefeng},
  journal={arXiv preprint arXiv:2311.17117},
  year={2023}
}

@inproceedings{goel2023humans,
  title={Humans in 4d: Reconstructing and tracking humans with transformers},
  author={Goel, Shubham and Pavlakos, Georgios and Rajasegaran, Jathushan and Kanazawa, Angjoo and Malik, Jitendra},
  booktitle=ICCV,
  pages={14783--14794},
  year={2023}
}

@inproceedings{shen2024world,
  title={World-grounded human motion recovery via gravity-view coordinates},
  author={Shen, Zehong and Pi, Huaijin and Xia, Yan and Cen, Zhi and Peng, Sida and Hu, Zechen and Bao, Hujun and Hu, Ruizhen and Zhou, Xiaowei},
  booktitle={SIGGRAPH Asia 2024 Conference Papers},
  pages={1--11},
  year={2024}
}

@article{chefer2025videojam,
  title={Videojam: Joint appearance-motion representations for enhanced motion generation in video models},
  author={Chefer, Hila and Singer, Uriel and Zohar, Amit and Kirstain, Yuval and Polyak, Adam and Taigman, Yaniv and Wolf, Lior and Sheynin, Shelly},
  journal={arXiv preprint arXiv:2502.02492},
  year={2025}
}

@inproceedings{khirodkar2024sapiens,
  title={Sapiens: Foundation for human vision models},
  author={Khirodkar, Rawal and Bagautdinov, Timur and Martinez, Julieta and Zhaoen, Su and James, Austin and Selednik, Peter and Anderson, Stuart and Saito, Shunsuke},
  booktitle=ECCV,
  pages={206--228},
  year={2024},
  organization={Springer}
}

@article{liu2023visual,
  title={Visual instruction tuning},
  author={Liu, Haotian and Li, Chunyuan and Wu, Qingyang and Lee, Yong Jae},
  journal=NIPS,
  volume={36},
  pages={34892--34916},
  year={2023}
}

@article{bai2025qwen2,
  title={Qwen2. 5-vl technical report},
  author={Bai, Shuai and Chen, Keqin and Liu, Xuejing and Wang, Jialin and Ge, Wenbin and Song, Sibo and Dang, Kai and Wang, Peng and Wang, Shijie and Tang, Jun and others},
  journal={arXiv preprint arXiv:2502.13923},
  year={2025}
}

@article{guo2025seed1,
  title={Seed1. 5-vl technical report},
  author={Guo, Dong and Wu, Faming and Zhu, Feida and Leng, Fuxing and Shi, Guang and Chen, Haobin and Fan, Haoqi and Wang, Jian and Jiang, Jianyu and Wang, Jiawei and others},
  journal={arXiv preprint arXiv:2505.07062},
  year={2025}
}

@article{brown2020language,
  title={Language models are few-shot learners},
  author={Brown, Tom and Mann, Benjamin and Ryder, Nick and Subbiah, Melanie and Kaplan, Jared D and Dhariwal, Prafulla and Neelakantan, Arvind and Shyam, Pranav and Sastry, Girish and Askell, Amanda and others},
  journal=NIPS,
  volume={33},
  pages={1877--1901},
  year={2020}
}

@article{agarwal2025cosmos,
  title={Cosmos world foundation model platform for physical ai},
  author={Agarwal, Niket and Ali, Arslan and Bala, Maciej and Balaji, Yogesh and Barker, Erik and Cai, Tiffany and Chattopadhyay, Prithvijit and Chen, Yongxin and Cui, Yin and Ding, Yifan and others},
  journal={arXiv preprint arXiv:2501.03575},
  year={2025}
}

@inproceedings{pavlakos2019expressive,
  title={Expressive body capture: 3d hands, face, and body from a single image},
  author={Pavlakos, Georgios and Choutas, Vasileios and Ghorbani, Nima and Bolkart, Timo and Osman, Ahmed AA and Tzionas, Dimitrios and Black, Michael J},
  booktitle=CVPR,
  pages={10975--10985},
  year={2019}
}

@inproceedings{siyao2024duolando,
  author    = {Siyao, Li and Gu, Tianpei and Yang, Zhitao and Lin, Zhengyu and Liu, Ziwei and Ding, Henghui and Yang, Lei and Loy, Chen Change},
  title     = {Duolando: Follower GPT with Off-Policy Reinforcement Learning for Dance Accompaniment},
  booktitle = {ICLR},
  year      = {2024},
}

@inproceedings{xu2024magicanimate,
  title={Magicanimate: Temporally consistent human image animation using diffusion model},
  author={Xu, Zhongcong and Zhang, Jianfeng and Liew, Jun Hao and Yan, Hanshu and Liu, Jia-Wei and Zhang, Chenxu and Feng, Jiashi and Shou, Mike Zheng},
  booktitle=CVPR,
  pages={1481--1490},
  year={2024}
}

@article{wang2024vividpose,
  title={Vividpose: Advancing stable video diffusion for realistic human image animation},
  author={Wang, Qilin and Jiang, Zhengkai and Xu, Chengming and Zhang, Jiangning and Wang, Yabiao and Zhang, Xinyi and Cao, Yun and Cao, Weijian and Wang, Chengjie and Fu, Yanwei},
  journal={arXiv preprint arXiv:2405.18156},
  year={2024}
}

@inproceedings{chang2025x,
  title={X-dyna: Expressive dynamic human image animation},
  author={Chang, Di and Xu, Hongyi and Xie, You and Gao, Yipeng and Kuang, Zhengfei and Cai, Shengqu and Zhang, Chenxu and Song, Guoxian and Wang, Chao and Shi, Yichun and others},
  booktitle=CVPR,
  pages={5499--5509},
  year={2025}
}

@inproceedings{kim2024tcan,
  title={Tcan: Animating human images with temporally consistent pose guidance using diffusion models},
  author={Kim, Jeongho and Kim, Min-Jung and Lee, Junsoo and Choo, Jaegul},
  booktitle=ECCV,
  pages={326--342},
  year={2024},
  organization={Springer}
}

@article{zhang2024mimicmotion,
  title={Mimicmotion: High-quality human motion video generation with confidence-aware pose guidance},
  author={Zhang, Yuang and Gu, Jiaxi and Wang, Li-Wen and Wang, Han and Cheng, Junqi and Zhu, Yuefeng and Zou, Fangyuan},
  journal={arXiv preprint arXiv:2406.19680},
  year={2024}
}

@article{shao2025interspatial,
  title={Interspatial Attention for Efficient 4D Human Video Generation},
  author={Shao, Ruizhi and Xu, Yinghao and Shen, Yujun and Yang, Ceyuan and Zheng, Yang and Chen, Changan and Liu, Yebin and Wetzstein, Gordon},
  journal={arXiv preprint arXiv:2505.15800},
  year={2025}
}

@inproceedings{karras2023dreampose,
  title={Dreampose: Fashion image-to-video synthesis via stable diffusion},
  author={Karras, Johanna and Holynski, Aleksander and Wang, Ting-Chun and Kemelmacher-Shlizerman, Ira},
  booktitle=ICCV,
  pages={22623--22633},
  year={2023},
  organization={IEEE}
}

@inproceedings{kanazawa2018end,
  title={End-to-end recovery of human shape and pose},
  author={Kanazawa, Angjoo and Black, Michael J and Jacobs, David W and Malik, Jitendra},
  booktitle=CVPR,
  pages={7122--7131},
  year={2018}
}

@inproceedings{rajasegaran2022tracking,
  title={Tracking people by predicting 3d appearance, location and pose},
  author={Rajasegaran, Jathushan and Pavlakos, Georgios and Kanazawa, Angjoo and Malik, Jitendra},
  booktitle=CVPR,
  pages={2740--2749},
  year={2022}
}

@inproceedings{kanazawa2019learning,
  title={Learning 3d human dynamics from video},
  author={Kanazawa, Angjoo and Zhang, Jason Y and Felsen, Panna and Malik, Jitendra},
  booktitle=CVPR,
  pages={5614--5623},
  year={2019}
}

@inproceedings{kocabas2020vibe,
  title={Vibe: Video inference for human body pose and shape estimation},
  author={Kocabas, Muhammed and Athanasiou, Nikos and Black, Michael J},
  booktitle=CVPR,
  pages={5253--5263},
  year={2020}
}

@inproceedings{luo20203d,
  title={3d human motion estimation via motion compression and refinement},
  author={Luo, Zhengyi and Golestaneh, S Alireza and Kitani, Kris M},
  booktitle=ACCV,
  year={2020}
}

@article{jiang2023motiongpt,
  title={Motiongpt: Human motion as a foreign language},
  author={Jiang, Biao and Chen, Xin and Liu, Wen and Yu, Jingyi and Yu, Gang and Chen, Tao},
  journal=NIPS,
  volume={36},
  pages={20067--20079},
  year={2023}
}

@inproceedings{jiang2025solami,
  title={Solami: Social vision-language-action modeling for immersive interaction with 3d autonomous characters},
  author={Jiang, Jianping and Xiao, Weiye and Lin, Zhengyu and Zhang, Huaizhong and Ren, Tianxiang and Gao, Yang and Lin, Zhiqian and Cai, Zhongang and Yang, Lei and Liu, Ziwei},
  booktitle=CVPR,
  pages={26887--26898},
  year={2025}
}

@article{xie2025show,
  title={Show-o2: Improved Native Unified Multimodal Models},
  author={Xie, Jinheng and Yang, Zhenheng and Shou, Mike Zheng},
  journal={arXiv preprint arXiv:2506.15564},
  year={2025}
}

@article{wang2024emu3,
  title={Emu3: Next-token prediction is all you need},
  author={Wang, Xinlong and Zhang, Xiaosong and Luo, Zhengxiong and Sun, Quan and Cui, Yufeng and Wang, Jinsheng and Zhang, Fan and Wang, Yueze and Li, Zhen and Yu, Qiying and others},
  journal={arXiv preprint arXiv:2409.18869},
  year={2024}
}

@article{ding2025mtvcrafter,
  title={MTVCrafter: 4D Motion Tokenization for Open-World Human Image Animation},
  author={Ding, Yanbo and Hu, Xirui and Guo, Zhizhi and Zhang, Chi and Wang, Yali},
  journal={arXiv preprint arXiv:2505.10238},
  year={2025}
}

@article{zheng2025vbench,
  title={Vbench-2.0: Advancing video generation benchmark suite for intrinsic faithfulness},
  author={Zheng, Dian and Huang, Ziqi and Liu, Hongbo and Zou, Kai and He, Yinan and Zhang, Fan and Zhang, Yuanhan and He, Jingwen and Zheng, Wei-Shi and Qiao, Yu and others},
  journal={arXiv preprint arXiv:2503.21755},
  year={2025}
}

@inproceedings{li2021ai,
  title={Ai choreographer: Music conditioned 3d dance generation with aist++},
  author={Li, Ruilong and Yang, Shan and Ross, David A and Kanazawa, Angjoo},
  booktitle=ICCV,
  pages={13401--13412},
  year={2021}
}

@inproceedings{shin2024wham,
  title={Wham: Reconstructing world-grounded humans with accurate 3d motion},
  author={Shin, Soyong and Kim, Juyong and Halilaj, Eni and Black, Michael J},
  booktitle=CVPR,
  pages={2070--2080},
  year={2024}
}

@inproceedings{cheng2023dna,
  title={Dna-rendering: A diverse neural actor repository for high-fidelity human-centric rendering},
  author={Cheng, Wei and Chen, Ruixiang and Fan, Siming and Yin, Wanqi and Chen, Keyu and Cai, Zhongang and Wang, Jingbo and Gao, Yang and Yu, Zhengming and Lin, Zhengyu and others},
  booktitle=ICCV,
  pages={19982--19993},
  year={2023}
}

@inproceedings{black2023bedlam,
  title={Bedlam: A synthetic dataset of bodies exhibiting detailed lifelike animated motion},
  author={Black, Michael J and Patel, Priyanka and Tesch, Joachim and Yang, Jinlong},
  booktitle=CVPR,
  pages={8726--8737},
  year={2023}
}

@inproceedings{vonMarcard2018,
    title = {Recovering Accurate 3D Human Pose in The Wild Using IMUs and a Moving Camera},
    author = {von Marcard, Timo and Henschel, Roberto and Black, Michael and Rosenhahn, Bodo and Pons-Moll, Gerard},
    booktitle = ECCV,
    year = {2018},
    month = {sep}
    }

@inproceedings{Huang:CVPR:2022,
title = {Capturing and Inferring Dense Full-Body Human-Scene Contact},
author = {Huang, Chun-Hao P. and Yi, Hongwei and H{\"o}schle, Markus and Safroshkin, Matvey and Alexiadis, Tsvetelina and Polikovsky, Senya and Scharstein, Daniel and Black, Michael J.},
  booktitle = CVPR,
  month = jun,
  year = {2022},
  pages = {13274-13285},
  month_numeric = {6}
}

@inproceedings{wu2025janus,
  title={Janus: Decoupling visual encoding for unified multimodal understanding and generation},
  author={Wu, Chengyue and Chen, Xiaokang and Wu, Zhiyu and Ma, Yiyang and Liu, Xingchao and Pan, Zizheng and Liu, Wen and Xie, Zhenda and Yu, Xingkai and Ruan, Chong and others},
  booktitle=CVPR,
  pages={12966--12977},
  year={2025}
}

@article{chen2025janus,
  title={Janus-pro: Unified multimodal understanding and generation with data and model scaling},
  author={Chen, Xiaokang and Wu, Zhiyu and Liu, Xingchao and Pan, Zizheng and Liu, Wen and Xie, Zhenda and Yu, Xingkai and Ruan, Chong},
  journal={arXiv preprint arXiv:2501.17811},
  year={2025}
}

@article{wu2024deepseek,
  title={Deepseek-vl2: Mixture-of-experts vision-language models for advanced multimodal understanding},
  author={Wu, Zhiyu and Chen, Xiaokang and Pan, Zizheng and Liu, Xingchao and Liu, Wen and Dai, Damai and Gao, Huazuo and Ma, Yiyang and Wu, Chengyue and Wang, Bingxuan and others},
  journal={arXiv preprint arXiv:2412.10302},
  year={2024}
}

@inproceedings{wang2025seedvr,
  title={Seedvr: Seeding infinity in diffusion transformer towards generic video restoration},
  author={Wang, Jianyi and Lin, Zhijie and Wei, Meng and Zhao, Yang and Yang, Ceyuan and Loy, Chen Change and Jiang, Lu},
  booktitle=CVPR,
  pages={2161--2172},
  year={2025}
}

@article{chen2024motionllm,
  title={Motionllm: Understanding human behaviors from human motions and videos},
  author={Chen, Ling-Hao and Lu, Shunlin and Zeng, Ailing and Zhang, Hao and Wang, Benyou and Zhang, Ruimao and Zhang, Lei},
  journal={arXiv preprint arXiv:2405.20340},
  year={2024}
}

@inproceedings{li2025human,
  title={Human motion instruction tuning},
  author={Li, Lei and Jia, Sen and Wang, Jianhao and Jiang, Zhongyu and Zhou, Feng and Dai, Ju and Zhang, Tianfang and Wu, Zongkai and Hwang, Jenq-Neng},
  booktitle=CVPR,
  pages={17582--17591},
  year={2025}
}

@article{zhu2025motiongpt3,
  title={MotionGPT3: Human Motion as a Second Modality},
  author={Zhu, Bingfan and Jiang, Biao and Wang, Sunyi and Tang, Shixiang and Chen, Tao and Luo, Linjie and Zheng, Youyi and Chen, Xin},
  journal={arXiv preprint arXiv:2506.24086},
  year={2025}
}

@article{luo2024m,
  title={M$^{3}$GPT: An Advanced Multimodal, Multitask Framework for Motion Comprehension and Generation},
  author={Luo, Mingshuang and Hou, Ruibing and Li, Zhuo and Chang, Hong and Liu, Zimo and Wang, Yaowei and Shan, Shiguang},
  journal=NIPS,
  volume={37},
  pages={28051--28077},
  year={2024}
}

@article{li2025chatmotion,
  title={Chatmotion: A multimodal multi-agent for human motion analysis},
  author={Li, Lei and Jia, Sen and Wang, Jianhao and An, Zhaochong and Li, Jiaang and Hwang, Jenq-Neng and Belongie, Serge},
  journal={arXiv preprint arXiv:2502.18180},
  year={2025}
}

@inproceedings{li2025unipose,
  title={Unipose: A unified multimodal framework for human pose comprehension, generation and editing},
  author={Li, Yiheng and Hou, Ruibing and Chang, Hong and Shan, Shiguang and Chen, Xilin},
  booktitle=CVPR,
  pages={27805--27815},
  year={2025}
}

@article{xi2025omnivdiff,
  title={Omnivdiff: Omni controllable video diffusion for generation and understanding},
  author={Xi, Dianbing and Wang, Jiepeng and Liang, Yuanzhi and Qiu, Xi and Huo, Yuchi and Wang, Rui and Zhang, Chi and Li, Xuelong},
  journal={arXiv preprint arXiv:2504.10825},
  year={2025}
}

@article{dang2025svimo,
  title={SViMo: Synchronized Diffusion for Video and Motion Generation in Hand-object Interaction Scenarios},
  author={Dang, Lingwei and Shao, Ruizhi and Zhang, Hongwen and Min, Wei and Liu, Yebin and Wu, Qingyao},
  journal={arXiv preprint arXiv:2506.02444},
  year={2025}
}

@article{huang2025animax,
  title={AnimaX: Animating the Inanimate in 3D with Joint Video-Pose Diffusion Models},
  author={Huang, Zehuan and Feng, Haoran and Sun, Yangtian and Guo, Yuanchen and Cao, Yanpei and Sheng, Lu},
  journal={arXiv preprint arXiv:2506.19851},
  year={2025}
}
}


\end{document}